\documentclass[letterpaper, 10 pt, conference]{ieeeconf}  

\IEEEoverridecommandlockouts                              

\usepackage{multirow}
\usepackage[hyphens]{url}
\usepackage{graphicx}
\usepackage{gensymb}
\usepackage{xcolor}

\title{\LARGE \bf
Towards Complex and Continuous Manipulation: A Gesture Based Anthropomorphic Robotic Hand Design}

\author{Li Tian, Hanhui Li, Qifa Wang, Xuezeng Du, Jialin Tao, Jordan Sia Chong, \\ Nadia Magnenat Thalmann, and Jianmin Zheng
\thanks{All the authors are with Nanyang Technological University, Singapore. This research is supported by the National Research Foundation, Singapore under its International Research Centres in Singapore Funding Initiative, and Institute for Media Innovation, Nanyang Technological University (IMI-NTU). Any opinions, findings and conclusions or recommendations expressed in this material are those of the author(s) and do not reflect the views of National Research Foundation, Singapore. Hanhui Li is the corresponding author (hanhui.li@ntu.edu.sg).}%
}
\begin{document}

\maketitle
\thispagestyle{empty}
\pagestyle{empty}

\begin{abstract}
Most current anthropomorphic robotic hands can realize part of the human hand functions, particularly for object grasping. However, due to the complexity of the human hand, few current designs target at daily object manipulations, even for simple actions like rotating a pen. To tackle this problem, we introduce a gesture based framework, which adopts the widely-used 33 grasping gestures of Feix as the bases for hand design and implementation of manipulation. In the proposed framework, we first measure the motion ranges of human fingers for each gesture, and based on the results, we propose a simple yet dexterous robotic hand design with 13 degrees of actuation. Furthermore, we adopt a frame interpolation based method, in which we consider the base gestures as the key frames to represent a manipulation task, and use the simple linear interpolation strategy to accomplish the manipulation. To demonstrate the effectiveness of our framework, we define a three-level benchmark, which includes not only 62 test gestures from previous research, but also multiple complex and continuous actions. Experimental results on this benchmark validate the dexterity of the proposed design and our video is available in \url{https://drive.google.com/file/d/1wPtkd2P0zolYSBW7_3tVMUHrZEeXLXgD/view?usp=sharing}. 

\end{abstract}

\section{Introduction}

\begin{figure*}[!t]
	\centering
	\includegraphics[width = 0.95\linewidth]{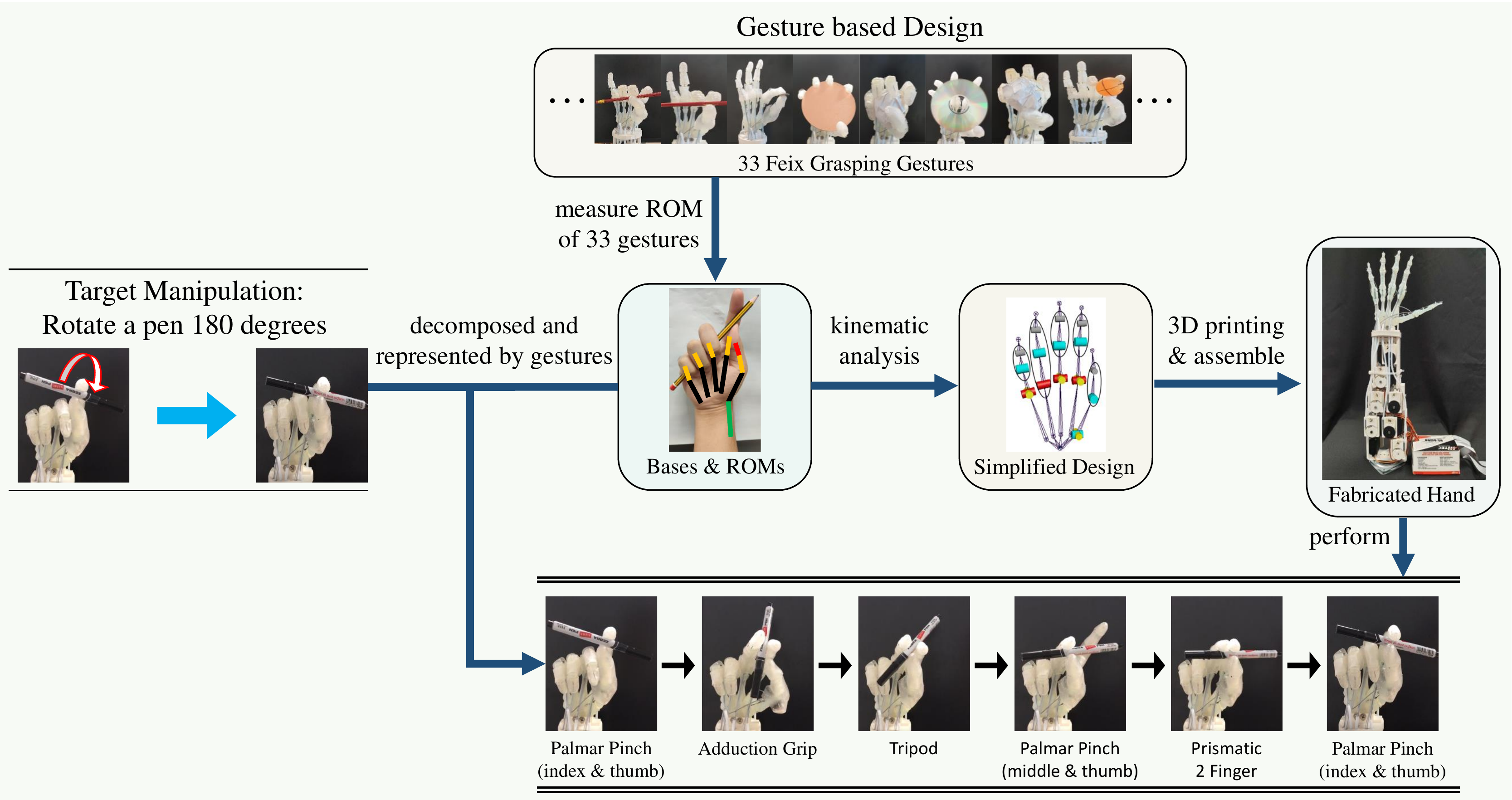}
	\caption{Demonstration of the proposed gesture based framework. We consider 33 Feix's Grasping gestures as the bases, and measure the ROM of the human hand for achieving each gesture. With the results, we design a simplified robotic hand with the necessary ROMs. Consequently, given a target manipulation, we can represent and complete it with multiple gestures.}
	\label{fig:framework}
\end{figure*}

Artificial hands remain one of the hardest problems in robotics \cite{piazza2019century,liu2008multisensory}, due to the lack of comprehensive understanding on the actuation and sensory systems of the human hand. Earlier studies \cite{robonaut,shadowhand,SolveCube} try to fully replicate functions of the human hand via complicated mechanical structures and actuation systems. Although these robotic hands have ranges of motion (ROM) and degrees of freedom (DOF) similar to those of human hands, or even have the ability to complete astounding manipulation tasks like solving a Rubik's cube \cite{SolveCube}, they are costly to fabricate and their dexterity can still be improved.

Therefore, recent robotic hand designs focus more on simplifying mechanical structures and realizing partial functions of human hands. To the best of our knowledge, there are two notable development directions for such a purpose. The first direction is based on the concept of anatomically correct robotic hand \cite{ACT,ACB,faudzi2017index,tasi2019design}, which imitates the critical biomechanical structures of the human hand, such as bones, joints and tendons. This strategy does lower the complexity of the designing process, yet maintaining the balance between dexterity and complexity remains a challenge. For instance, in one of the latest anatomically correct robotic hand \cite{tasi2019design}, 30 servo motors are used to realize 33 grasping gestures of Feix \cite{feix2015grasp} and the Kapandji test \cite{KapandjiTest}.  

The other trend includes the soft robotic hands \cite{rus2015design} that utilize soft materials and actuators, such as gas/liquid pressure \cite{deimel2016novel,polygerinos2015soft,zhou2018bcl} and artificial muscles \cite{laschi2012soft,kurumaya2017design,diteesawat2020characteristic}. Robotic hands of this type have remarkable deformability, and hence they can accomplish the tasks of object grasping safely \cite{spiers2018variable,zhou2019soft,lu2019soft,abondance2020dexterous}. However, their mechanical structures and kinematic models are different from that of the human hand significantly. Hence it is difficult for them to realize natural and human-like manipulations.

Besides, underactuated designs that have fewer actuators than DOFs \cite{piazza2019century,birglen2007underactuated} are the popular choices for simplified robotic hands. For example, Gosselin et al. \cite{gosselin2008anthropomorphic} propose an anthropomorphic design of 15 DOFs with a single actuator. Zisimatos et al. \cite{zisimatos2014open} propose a modular underactuated design, which can be used to fabricate hands with different forms (gripper-like, two fingered, multi-fingered). Designs of this type allow the passive movements between DOFs \cite{piazza2019century}, at the cost of losing the full control of joints.

In our opinion, for the problem of object grasping, the performance of current methods is promising. Nevertheless, few current methods can tackle the problem of daily object manipulations, even for a simple action like spinning a pen. This is because they are highly specialized for their particular purposes and lose the adaptability. For example, a soft robotic hand \cite{zhou2018bcl} targeting at object grasping might not realize the functions of abduction and adduction for all fingers. This problem should not be ignored, as manipulating common objects in a human-like way, is one of the basic requirements for social robots and humanoids.

To address the above issues, we propose a novel framework, in which the design of robotic hands and object manipulations depend on a set of grasping gestures. The intuition behind our framework is that, most daily object manipulation tasks can be represented by the set of grasping gestures. Such a strategy is possible, as the pioneer research by Bullock et al. \cite{bullock2012hand} has tried to classify common manipulation tasks at a coarse level. Meanwhile, we can measure the ROM of each base gesture, so that our robotic hand design can be simplified to provide the only necessary ROMs. In this way, the proposed design can achieve the excellent balance between simplicity and dexterity. Besides, since there are extensive studies on achieving various grasping gestures, we can avoid building our solution from scratch, while current methods can be augmented with our framework to tackle the object manipulation problem.

Consequently, based on the proposed framework, we design an anthropomorphic hand with 13 degrees of actuation (DOA), which is much simpler yet more versatile compared with previous methods \cite{tasi2019design,takamuku2008repetitive}. Besides, we propose a key-frame interpolation method, in which we select multiple base gestures as the key frames, and then adopt the simple linear interpolation to control the process of manipulating an object. To fully evaluate the dexterity of our design, we further introduce the \emph{complex and continuous manipulation} benchmark, which has tasks of three difficulty levels. 

In summary, this paper makes noteworthy contributions to anthropomorphic robotic hand design as follows:

\textbullet We introduce a novel framework to provide previous grasping-oriented methods with the adaptability to object manipulations, which also sheds light on designing highly dexterous robotic hands.

\textbullet We collect ROMs of a set of base gestures, which are valuable data for robotic hand design and algorithm development for object manipulation.

\textbullet We propose a simplified dexterous hand model of 13 DOAs and a key-frame interpolation method, which are validated on our three-level benchmark for object grasping and manipulation.

\section{The gesture based framework}

\begin{table}[!t]
	\centering
	\caption{Terms and abbreviations used in this paper.}
	\label{tab: abbr}
	\begin{tabular}[c]{|*{2}{c|}}
		\hline
		\textbf{Abbreviation} & \textbf{Definition} \\ \hline
		Abd/Add & abduction/adduction \\ \hline
		DIP  & Distal interphalangeal \\ \hline
		PIP  & Proximal interphalangeal \\ \hline
		MCP  & Metacarpal \\ \hline
		CMC  & Carpometacarpal \\ \hline
		LUM  & Lumbrical muscles \\ \hline
		DOA  & Degree of actuation \\ \hline
		DOF  & Degree of freedom \\ \hline
		ROM  & Range of motion \\ \hline
		CCM  & Complex and continuous manipulation \\ \hline
	\end{tabular}
\end{table}


In this section, we introduce the proposed gesture based framework in detail. As demonstrated in Fig. \ref{fig:framework}, we first select a set of grasping gestures as the bases. We then measure the ROM of the human hand for each gesture, so that we can design a robotic hand with the necessary dexterity to accomplish the base gestures. Finally, given a manipulation task, we choose multiple gestures as the key frames, and adopt the interpolation method to control our robotic hand.

Due to the complexity of the human hand anatomy, we summarize the terms and abbreviations related to the human hand structures in Table \ref{tab: abbr} for easy understanding and reading.

\begin{figure}[!t]
	\centering
	\includegraphics[width = 0.95\linewidth]{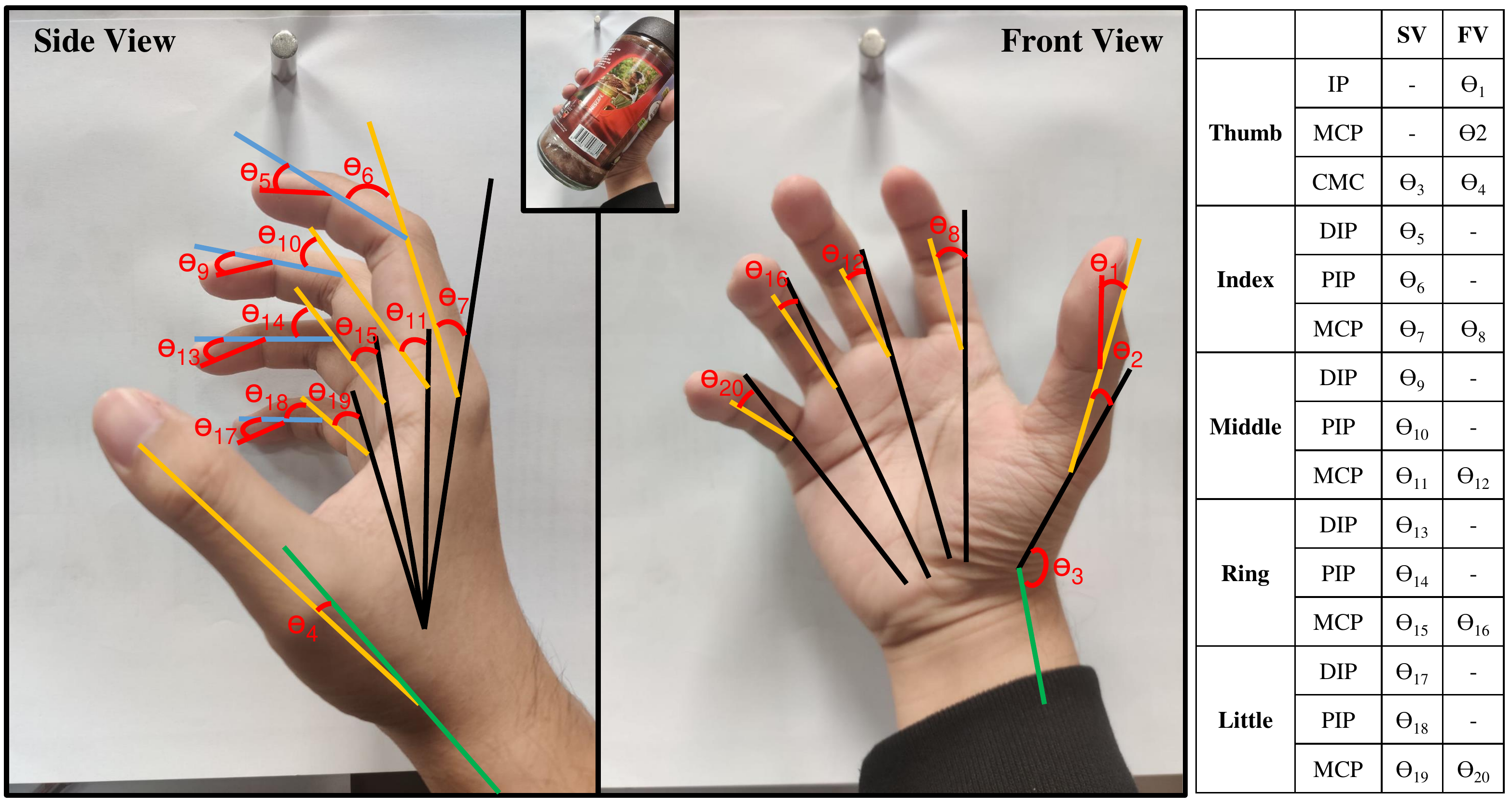}
	\caption{Measuring 20 angles of joint rotation for an exemplar gesture. Best viewed on a high-resolution screen.}
	\label{fig:angleMeasure}
\end{figure}

\subsection{The Base Gestures}

Selecting the base gestures allows us to understand the essential functions that need to be implemented. In this paper, we choose the grasp taxonomy defined by Feix et al. \cite{feix2015grasp}, which has organized the common human hand configurations into 33 gestures. Any other gesture, such as those defined in \cite{bullock2012hand,benchmark50}, can be included but we find that these 33 gestures are adequate for various in-hand manipulation tasks. 

Given the selected base gestures, we consider the human hand as the template for measuring the ROMs. Previous studies \cite{salisbury1982articulated,zhou2018bcl} suggest that each finger can be characterized with a kinematic model of 3 DOFs (3 joints with flexion and extension only). Hence, for each gesture, we annotate 15 landmarks on the joints manually from two views. We use the angles of joint rotation as the metric of ROM. In order to obtain a precise template, besides flexion and extension, we also take abduction and adduction into account, and hence 4 joint rotation angles will be recorded for each digit. An example of measuring all 20 angles is shown in Fig. \ref{fig:angleMeasure}. Each gesture is measured 3 times, and we use the average rotation angles as the final representation of the gesture\footnote{The ROMs of our 33 gestures can be found in \url{https://drive.google.com/file/d/1NLzuUmtjXN2oWuPHLAG26-SRDqi_308f/view?usp=sharing}}.

\begin{figure}[!t]
	\centering
	\includegraphics[width = 0.95\linewidth]{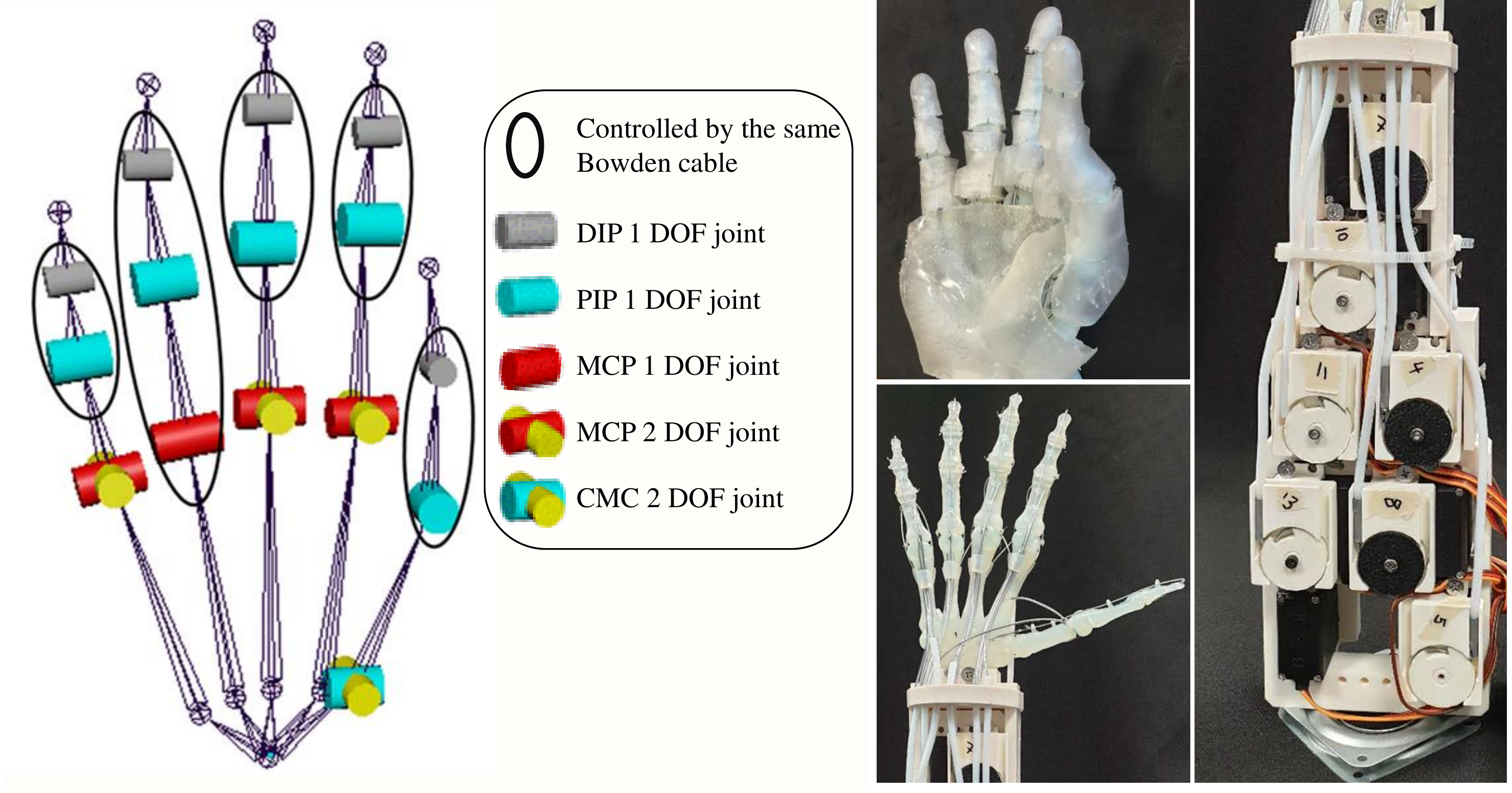}
	\caption{The 13-DOA kinematic model (left) and the fabricated hand of our gesture based design (right).}
	\label{fig:fabricated}
\end{figure}

\subsection{Design of The Proposed Hand}
The 33 base gestures provide the foundation of the kinematic design of our robotic hand. Particularly, our goals are to devise a simplified hand that can complete these base gestures, while lower its complexity by reducing the number of actuators as many as possible.

To begin with, it is noted that the distal interphalangeal (DIP) and proximal interphalangeal (PIP) joints are always bent together \cite{lin2000modeling}. Therefore, the DIP joint and the PIP joint of each finger\footnote{Strictly speaking, these two joints of the thumb are called the interphalangeal joint and the metacarpal joint, yet for the convenience of expression, we still consider them as DIP and PIP in this paper.} can be controlled by a single actuator. This is achieved by the active control of flexion/extension of the DIP joint, along with the passive movements of the PIP joint. Secondly, for the metacarpal (MCP) joint and carpometacarpal (CMC) joint, they have 2 DOFs for flexion/extension and abduction/adduction, respectively. Based on our measured data, most of the 33 gestures requires these 2 DOFs for the thumb, the index finger and the middle finger. Therefore, for each of these three fingers, we utilize an actuator for flexion/extension and another one for abduction/adduction. As to the ring finger and the little finger, these 2 DOFs are less important and hence we consider to remove them from one of these two fingers. We notice that the ROM of the MCP joint of the little finger for abduction/adduction ($[-14, 20]$) is larger than that of the ring finger ($[-6, 18]$). Furthermore, the MCP's function of the little finger is important for holding the objects in several gestures, such as ``Prismatic 4 Finger", ``Precision Disk", and ``Precision Sphere". Therefore, we drive the little finger with three actuators as in other three fingers, while simplify the structure of the ring finger by controlling its three joints with the same actuator. In this way, we construct a simplified underactuated robotic hand of 13 DOAs, as demonstrated in Fig. \ref{fig:design}. 

\begin{table*}[!t]
	\centering
	\caption{Comparison of dexterous hand designs. Our mechanical structure is simpler yet able to complete more gestures.}
	\label{tab: design_comp}
	\begin{tabular}[c]{|c|c|c|c|}
		\hline
		& \textbf{Biomimetic Hand} \cite{ACB} & \textbf{HR-hand} \cite{faudzi2017index}& \textbf{Ours} \\ \hline
		Bone & 3D printed, ABS & Molded, resin & 3D printed, resin \\ \hline
		
		Actuator & Servo: $9 \times$ MX-12W, $1 \times$ AX-12A & Pneumatic McKibben actuators & Servo: $13 \times$ MG996R \\ \hline
		
		Tendon and Muscle & FDP, EDC & FDP, FDS, EDC, LUM, INT& FDP, LUM INT \\ \hline
		
		Driven Cable& Soft cable & Soft cable & Bowden cable  \\ \hline
		
		Ligament & Crocheted ligament & Silicone ligament & Fishing wire \\ \hline
		
		Tendon Pulley & Laser-cut elastic silicone & Polyethylene tubes & 3D printed with bones\\ \hline
		
		Tissue and Skin & Fingertip cap only & N.A. & Elastic tissue, silicone skin \\ \hline
		Evaluation Method & 15 gestures & ROM of finger & 62 gestures and CCM \\ \hline
	\end{tabular}
\end{table*}

As we will report in the experimental section, our design can complete the base gestures robustly. Note that we do not assume that our design is the only feasible one. For example, we can also simplify the structure of the little finger, with the risk of degraded performance. Nevertheless, our design has preserved the advantages of anatomically correct robotic hands in resembling the biomechanical structures of the human hand. Moreover, as it needs to meet the minimum ROM requirements of the base grasping gestures, the dexterity of our fabricated hand is guaranteed.

\subsection{Materials and Fabrication}

\begin{figure}[!t]
	\centering
	\includegraphics[width = 1.0\linewidth]{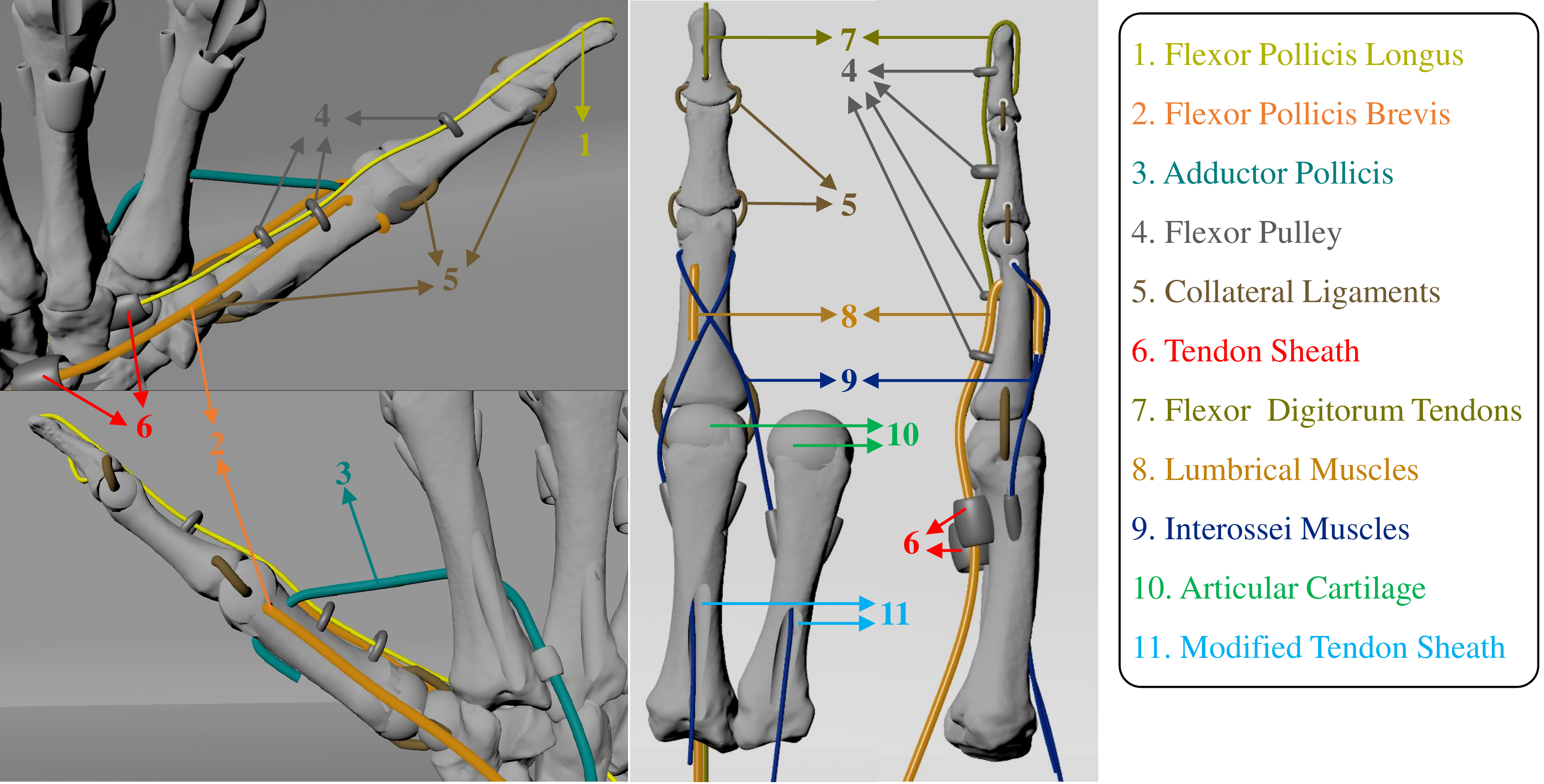}
	\caption{Examples of the mechanical structures of our fingers, including the thumb (left) and the index finger (right).}
	\label{fig:design}
\end{figure}

Our design can be fabricated efficiently by leveraging the advantages of 3D printing in fast manufacturing. Based on our previous studies on 3D modeling of hands \cite{tian2018methodology,tian2020fast}, we construct a printable 3D model consisting of three layers, i.e., the skin, tissues, and bones. Note that the tissue layer is a unique soft deformable structure, so that the 3D model can complete various grasping gestures successfully.  

The mechanical structures of the 3D model is shown in Fig. \ref{fig:design}. As mentioned above, Bowden cables are used as tendons and muscles to drive the phalanges: For the thumb, the flexor pollicis longus is used to control the flexion and extension of the DIP joint and the PIP joint, while the flexor pollicis brevis for that of the CMC joint. And the adductor pollicis is used for the abduction and adduction of the thumb. For the index finger, the middle finger and the little finger, the flexor digitorum tendons are used for the flexion and extension of the DIP joints and the PIP joints, while the lumbrical muscles (LUM) for that of the MCP joints. The abduction and adduction motions of these finger are controlled by the interossei muscles. The structure of the ring finger is omitted, as all its three joints are controlled by the same Bowden cable. All the Bowden cables we used in this paper are of $7$ strands with a diameter of 0.75 mm, except for those mimicking the LUM, which are of $19$ strands with a diameter of 1.25 mm and have a higher modulus of rigidity. 

Besides, the flexor pulleys are printed together with the bones to maintain the apposition of tendons and bones. Two types of plastic tendon sheaths are inserted into the bones to aid the routing of tendons. As in our previous design \cite{tian2020fast}, the collateral ligaments are necessary for connecting the bones. We use fishing wires as the collateral ligaments, which has a diameter of 0.34 mm, straight tension of 17 lb and knot tension of 15.5 lb. Articular cartilages of sphere shape are used to reduce the friction between bones.

The ``Form 2" 3D printer from Formlabs is used to fabricate the hand. As to the 3D printing materials, we use the same materials as in \cite{tian2020fast}, namely, \emph{rs-f2-gpcl-04} for bones and \emph{rs-f2-elcl-01} for the tissue layer. Our actuation system is composed of 13 MG996R servo motors (torque: 11 kg/cm). In Table \ref{tab: design_comp}, we summarize the configurations of our robotic hand, and compared it with two state-of-the-art methods. From this table, we can see that the complexity of our robotic hand is moderate, and as we will demonstrate in the experimental section, the proposed design is robust enough to handle different grasping and manipulation tasks. 

\subsection{Key-frame Interpolation for Object Manipulation}

Last but not least, we introduce a simple key-frame interpolation method to demonstrate the potential of the gesture based framework in tackling daily manipulation tasks. More advanced techniques, such as deep reinforcement learning \cite{SolveCube}, can be applied within the proposed framework, but this is beyond the scope of this paper. 

Given a target manipulation, e.g., rotating a pen 180 degrees, we can first represent it by multiple base gestures, which we refer as the key frames. These key frames can be selected manually, or via heuristic strategies, e.g., we can consider gestures as states in a Markov chain and learn to sample optimal states \cite{alterovitz2007stochastic}. Here we assume both the initial gesture and the end gesture are palmar pinch (with the index finger and the thumb), and the 4 selected key frames are adduction grip, tripod, palmar pinch (with the middle finger and the thumb), and prismatic 2 finger, as demonstrated in Fig. \ref{fig:framework}. Without loss of generality, let $\mathbf{\theta}_i$ denote the 20 ROM values of one of the key frames, and $\mathbf{\theta}_{i+1}$ of the next key frame. Then, given an interval of $T$ frames, we consider the simple linear interpolation method to obtain the ROM of the intermediate gesture at the $t$-th frame as follows:
\begin{equation}
	\mathbf{\theta}_{i,t} = \mathbf{\theta}_i + \frac{t({\mathbf{\theta}_{i + 1} - \mathbf{\theta}_i})}{T},
\end{equation}
where $1 \le t \le T$. Such a method is practical due to the following two reasons: First, the 33 base gestures of Feix are static and stable per se, which indicates that if the interval or ${\mathbf{\theta}_{i + 1} - \mathbf{\theta}_i}$ is small enough, the object can be considered to be relatively stable to the robotic hand. Second, within the selection process of the key frames, we can impose heuristic constrains, e.g., the motion of the robotic hand should not be obstructed by the object. In this way, we can decompose various daily manipulation tasks into gestures and perform them with our robotic hand.

\subsection{Limitations}
One of the limitations of our robotic hand is that, as in other 3D printed anthropomorphic hands \cite{ACB,faudzi2017index}, the ligaments for connecting bones are imitated by soft, non-grid materials. Consequently, robotic hands of this kind are less durable and weaker in strength, compared with those built with rigid structures \cite{shadowhand}. Besides, we have not applied any sensor into our robotic hand, and thus we cannot receive feedback in our control system.  

\begin{figure*}[!t]
	\centering
	\includegraphics[width = 1.0\textwidth]{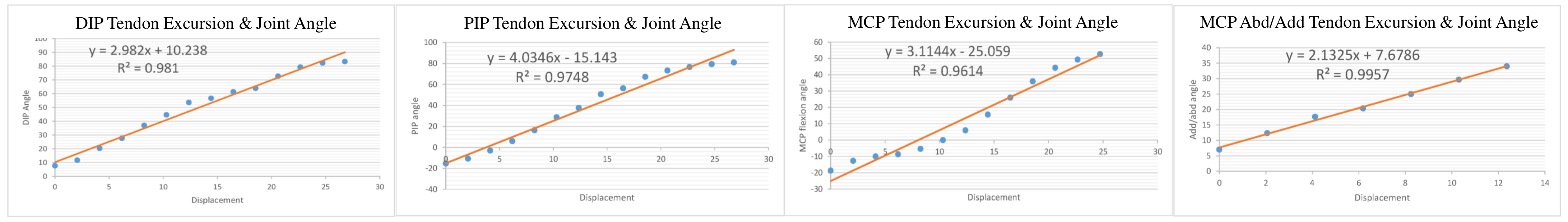}
	\caption{The relationship between tendon excursion (mm) and joint rotation angle (degree) of the index finger. Best viewed on a high-resolution screen.}
	\label{fig:curve}
\end{figure*}

\section{Experiment}
To validate the dexterity and quality of our gesture based robotic hand, we conduct extensive experiments in this paper. We first measuring multiple kinematic quantities of our robotic hand, and then perform qualitative analysis on a three-level benchmark for object manipulation.

\begin{figure}[!t]
	\centering
	\includegraphics[width = 0.75\linewidth]{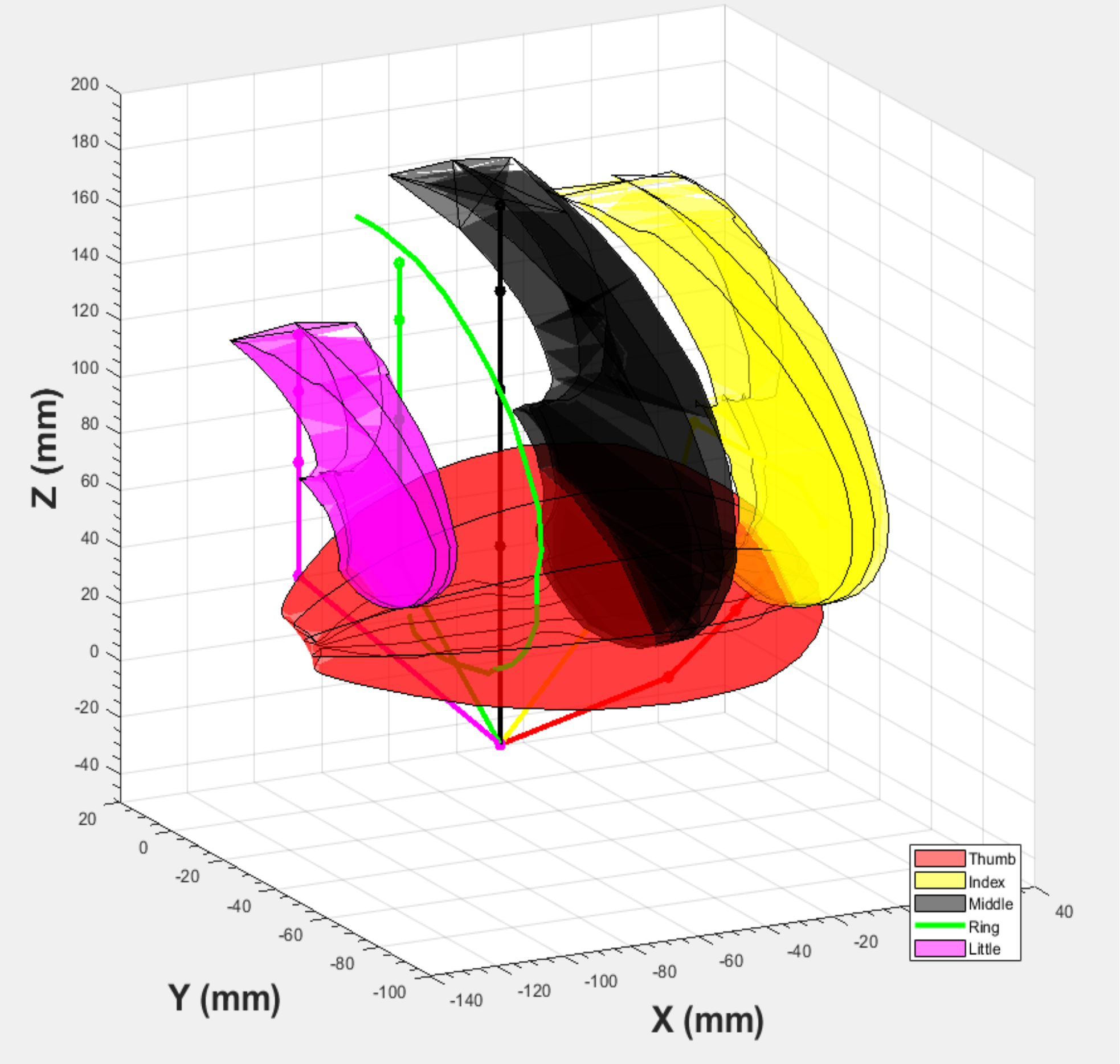}
	\caption{Fingertip trajectories of our robotic hand.}
	\label{fig:rom}
\end{figure}

\begin{figure}[!t]
	\centering
	\includegraphics[width = 1.0 \linewidth]{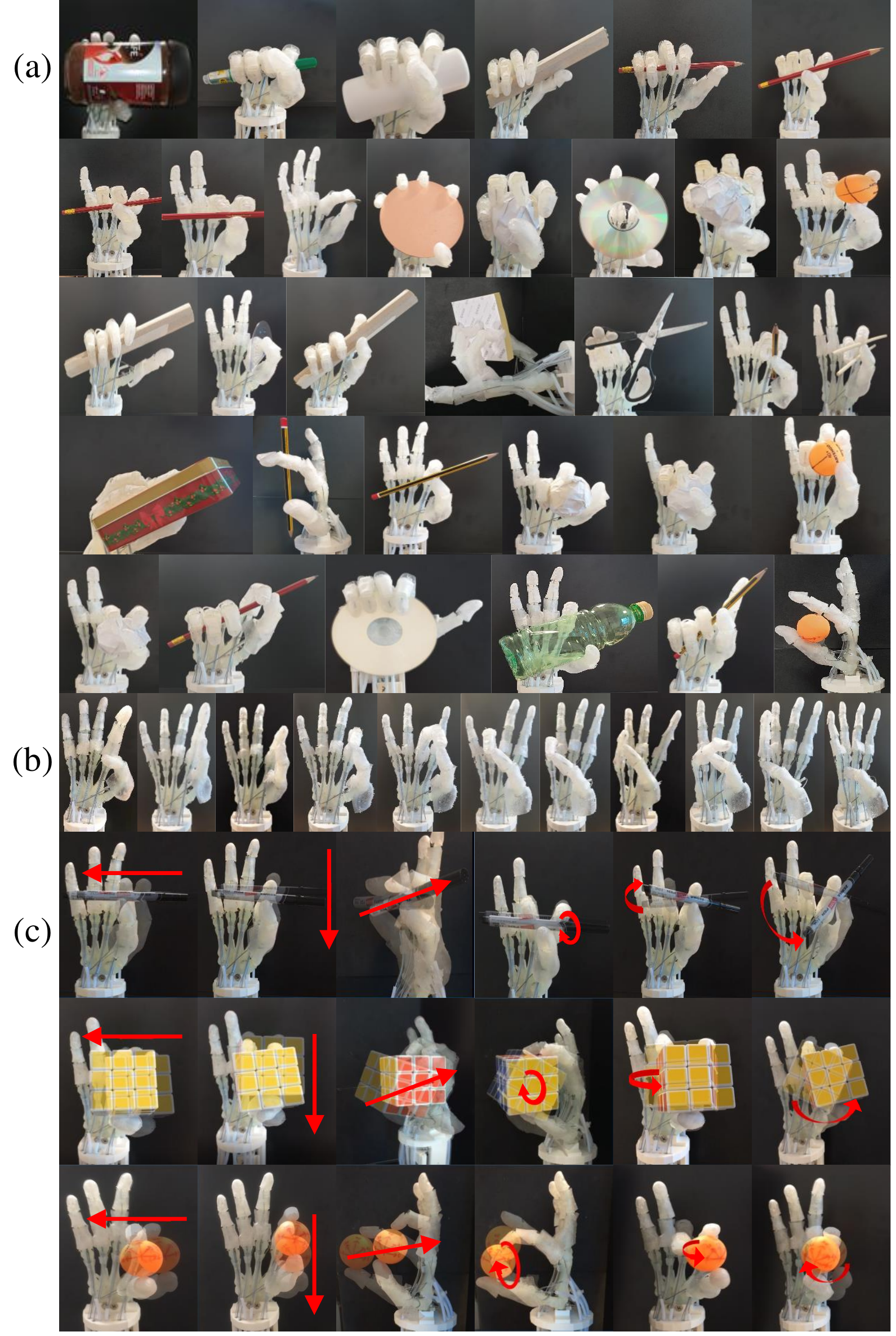}
	\caption{62 test gestures achieved by the proposed design: (a) 33 grasping gestures of Feix. (b) 11 Kapandji scores. (c) Translations and Rotations along the x, y, z axis, with 3 objects of different shapes (cylinder, cuboid and sphere). Best viewed on a high-resolution screen.}
	\label{fig:gesture}
\end{figure}

\begin{figure}[!t]
	\centering
	\includegraphics[width = 1.0 \linewidth]{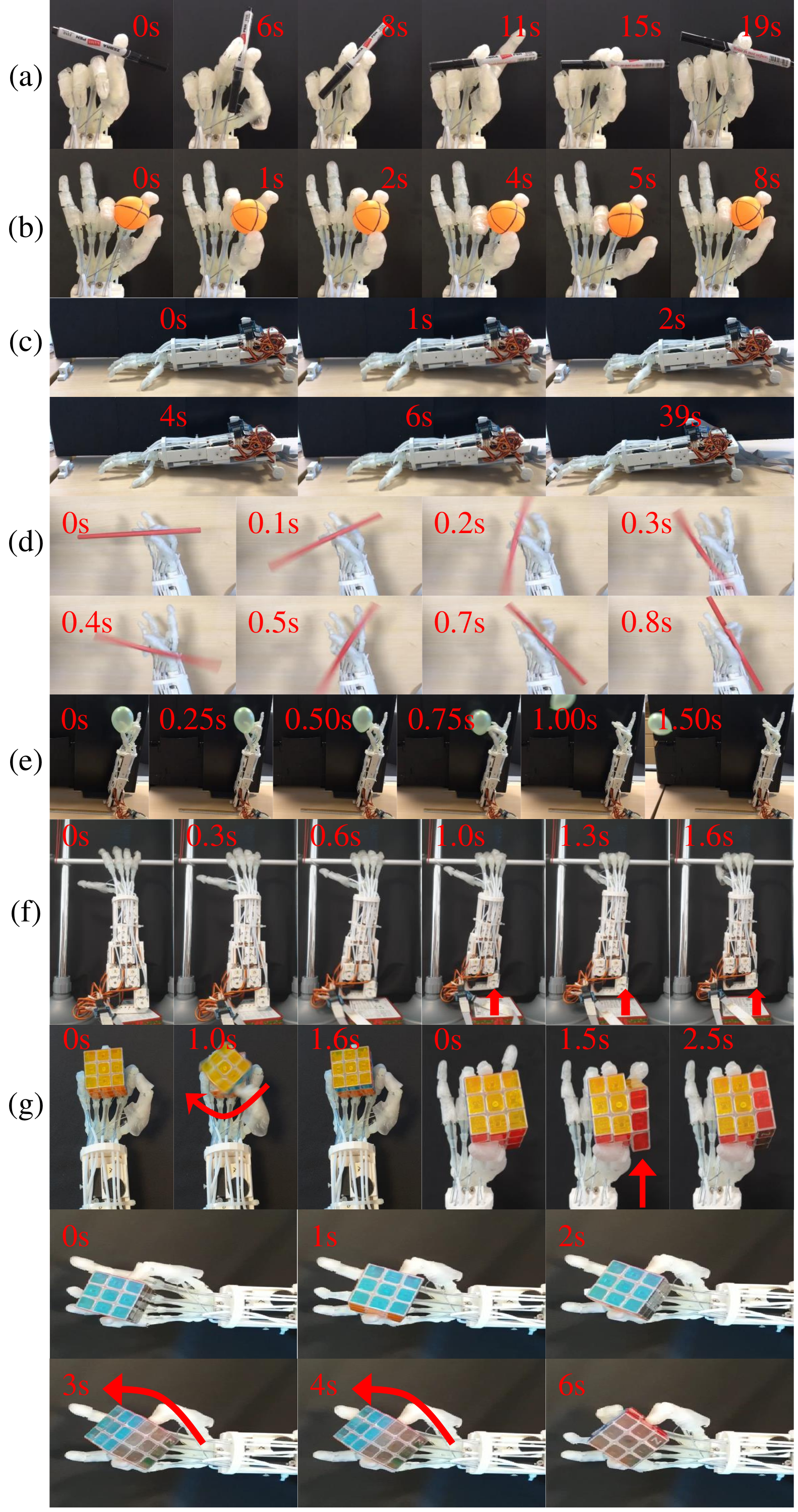}
	\caption{CCM completed by the proposed design. (a) Palmar pinching and rotating a pen. (b) Tripod based rotation of a ping-pong ball. (c) Crawling. (d) flicking and spinning a pen. (e) flicking a balloon. (f) climbing up. (g) rotating the top layer and right layer of a Rubik's cube, as well as overturning it. Best viewed on a high-resolution screen.}
	\label{fig:CCMResult}
\end{figure}

\subsection{Kinematics Analysis}
We first measure the relationship between tendon excursion and joint rotation angle of our robotic hand. As suggested in \cite{an1983tendon}, the tendon excursion $E$ and the joint angle $\phi$ of the human finger can be described as the following linear model:
\begin{equation}
	E = r\phi,
\end{equation}
where $r$ is the instantaneous moment arm. This indicates that if the relationship between $E$ and $\phi$ of our robotic hand also follows a similar linear function, then our design can replicate the movements of tendons precisely. Hence we record these data of our fabricated index finger for evaluation. The lengths of the distal, middle and proximal phalanges of the index finger are 25 mm, 25 mm and 43 mm, respectively. The results are reported in Fig. \ref{fig:curve} and they demonstrate the obvious liner pattern. We further fit these data with the linear regression model (i.e., $\phi = a E + b$), and obtain the coefficient of determination $R^2$ for all three joints. All $R^2$ values are larger than 0.96, which suggest that the linear model fits these data well and hence validate our design.

Following \cite{ACB}, we adopt the fingertip trajectory as a measure of motions. As presented in Fig. \ref{fig:rom}, the fingertip trajectories of our robotic hand resemble the logarithmic spiral curves, which are similar to those of the human hand \cite{ACB,kamper2003stereotypical}. We summarize the ROMs of the human hand \footnote{data from \url{https://www.verywellhealth.com/what-is-normal-range-of-motion-in-a-joint-3120361} and \url{https://www.orthopaedicsone.com/x/ygPbB}} and our robotic hand, as well as our measured ROMs of completing 33 grasping gestures in Table \ref{tab: rom}. Our design can duplicate about $83.8\%$ of the human hand ROMs and $90.1\%$ of the grasping ROMs, which guarantees our probability of tackling manipulation tasks. The maximum force generated from each fingertip are reported in Table \ref{tab: rom} as well. These results suggest that our design has the notable flexibility and the potential for completing various gestures.

\begin{table}[!t]
	\centering
	\scriptsize
	\caption{Comparison of the ROM of human hand and our robotic hand, and that for completing 33 grasping gestures. The length of each phalanx (in mm) and the maximum force (in N) generated from the fingertip of our robotic are reported as well.}
	\label{tab: rom}
	\begin{tabular}[c]{|*{7}{c|}}
		\hline
		\multirow{2}{*}{{\textbf{Finger}}} & \multirow{2}{*}{{\textbf{Joint}}} & \multirow{2}{*}{{\textbf{Human}}} & \multirow{2}{*}{{\textbf{Grasping}}} & \multicolumn{3}{c|}{\textbf{Ours}} \\ \cline{5-7} & & & & ROM & Length & Force \\ \hline
		\multirow{4}{*}{{Thumb}} & IP  & [0, 90] & [0, 84] & [0, 70] & 30 & \multirow{4}{*}{{10.8}} \\
		& MCP & [0, 70] & [0, 70] & [0, 75] & 32 & \\
		& CMC & [0, 53] & [0, 48] & [0, 61] & 49 & \\
		& Abd/Add & [-40, 50] & [0, 50] & [0, 50] & - & \\ \hline
		
		\multirow{4}{*}{{Index}} & DIP & [0, 80] & [0, 70] & [0, 70] & 25 & \multirow{4}{*}{{23.5}}\\
		& PIP & [0, 120]& [0, 100]& [0, 100] & 25 & \\
		& MCP & [0, 90] & [0, 90] & [-15, 82] & 43 & \\
		& Abd/Add & [-20, 25] & [-20, 22] & [-8, 26] & - & \\ \hline
		
		\multirow{4}{*}{{Middle}}& DIP & [0, 80] & [0, 80] & [0, 90] & 27 & \multirow{4}{*}{{21.6}} \\
		& PIP & [0, 120]& [0, 106]& [0, 80] & 31 & \\
		& MCP & [0, 90] & [0, 90] & [-7, 95] & 44 & \\
		& Abd/Add & [-20, 25] & [-20, 20] & [-6, 15] & - & \\ \hline
		
		\multirow{4}{*}{{Ring}}  & DIP & [0, 80] & [0, 73] & [0, 70] & 26 & \multirow{4}{*}{{22.6}}  \\
		& PIP & [0, 120]& [0, 120]& [0, 90] & 27 &\\
		& MCP & [0, 90] & [0, 90] & [-5, 95] & 46 & \\
		& Abd/Add & [-20, 25] & [-18, 6] & - & - & \\ \hline
		
		\multirow{4}{*}{{Little}}& DIP & [0, 80] & [0, 80] & [0, 69] & 25 & \multirow{4}{*}{{20.6}}\\
		& PIP & [0, 120]& [0, 110]& [0, 100] & 26 & \\
		& MCP & [0, 90] & [0, 90] & [-4, 90] & 27 & \\
		& Abd/Add & [-20, 25] & [-14, 20] & [-9, 28] & - & \\ \hline						
	\end{tabular}
\end{table}

\subsection{Complex and Continuous Manipulation}

We perform a series of object grasping and manipulation tasks to demonstrate the effectiveness of the proposed framework. Unlike conventional methods that only utilize single benchmark (e.g., Fexi's grasping gestures) for evaluation, here we introduce the idea of \textbf{complex and continuous manipulation}, which organizes daily manipulation tasks into the following three difficult levels:

\emph{Level 1: Single Grasping/Manipulation}. Tasks at this level can be tackled by a single gesture, such as the grasping gestures of Feix. As far as we know, most current methods are tested at this level.

\emph{Level 2: Complex Manipulation}. Tasks at this level require more than one gesture. For example, in our mentioned task of rotating a pen, 5 gestures are used in total. The key of completing tasks at this level is to ensure the high reproducibility of the base gestures and the stable transfer between gestures. 

\emph{Level 3: Complex and Continuous Manipulation} (CCM). For tasks at this level, we further impose the continuous constraint that, the manipulation must be completed within a limited time or with limited gestures. With this constraint, the mechanical design and the manipulation strategy must be optimized to remove redundant structures and gestures. We believe that CCM is one of the long-term targets of robotic hand research.

\begin{figure*}[!t]
	\centering
	\includegraphics[width = 0.75 \linewidth]{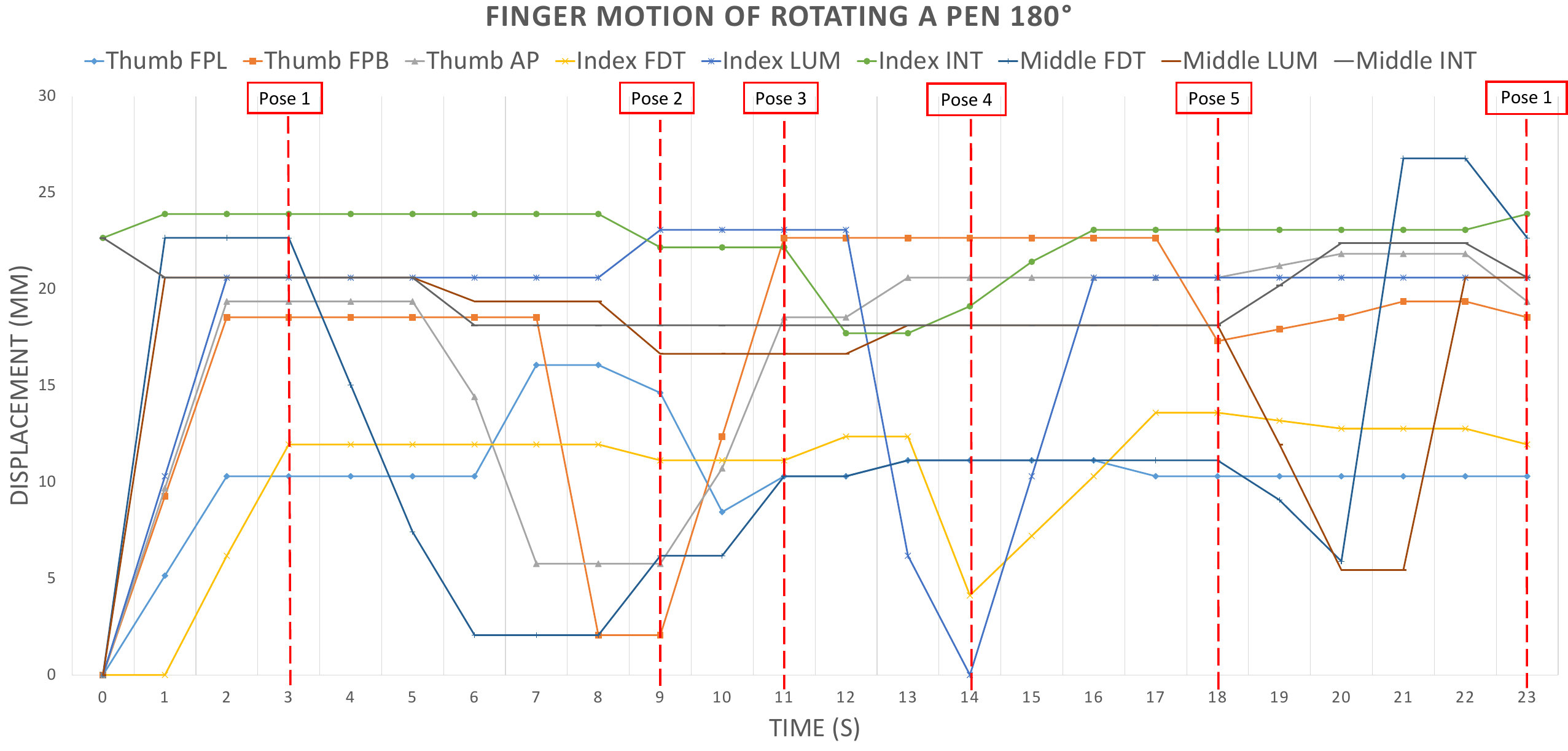}
	\caption{Displacements of joints during rotating a pen. Pose 1-5 are Palmar Pinch, Adduction Grip, Tripod, Palmar Pinch and Prismatic 2 Finger.}
	\label{fig:displacement}
\end{figure*}

As to the proposed methods, tasks of Level 1 can be completed successfully. We select 3 widely used benchmarks, 62 gesture in total, for testing our robotic hand, including 33 grasping gestures of Feix, 11 Kapandji scores \cite{KapandjiTest}, and 3 objects of different shapes (cylinder, cuboid and sphere) for translation and rotation test \cite{benchmark50}, as demonstrated in Fig. \ref{fig:gesture}. Each test is performed 10 times and the average success rate is larger than $93\%$. This result has built the solid background for us to handle tasks of level 2 and 3.

Due to the fact that there is no standard benchmark for object manipulation, we propose 7 tasks that cover as many base gestures as possible. As demonstrated in Fig. \ref{fig:CCMResult}, these tasks are (a) Palmar pinching and rotating a pen; (b) Tripod based rotation of a ping-pong ball along arbitrary direction; (c) Crawling; (d) flicking and spinning a pen; (e) flicking a balloon; (f) climbing up;  and (g) rotating the top layer and right layer of a Rubik's cube, as well as overturning the cube. With the proposed gesture based framework, we are able to complete all these tasks. Interested readers can refer to the supplemental video for more details. 

At last, we record the time and gestures for completing each of the 7 tasks, which could serve as the level 3 baseline for future research. As demonstrated in Fig. \ref{fig:CCMResult}, most of these tasks are completed within a few seconds. It is noted that the task of rotating a pen consumes 23 seconds, as demonstrated in Fig. \ref{fig:displacement}. This is because we use 5 gestures as the key frames. It is expected that with more advanced learning techniques, fewer gestures might be used and the processing time will be further reduced. Besides, our robotic also successfully performs multiple daily manipulation tasks, such as picking a pen and loosing a screw with a screwdriver. Interested readers can refer to our supplemental video for more details.

\section{Conclusion}

In this paper,  we propose the gesture based framework, of which the core is the set of base gestures and their ROMs. Given the widely-used gestures as the bases, we devise the simplified 13-DOA robotic hand to provide the essential functions for achieving them. We also demonstrate that complex manipulation tasks can be subdivided into multiple base gestures, and we adopt the key-frame interpolation method to complete the tasks. The proposed hand  is evaluated on our CCM benchmark and demonstrates its remarkable dexterity. In all, we hope that the gesture based framework, as well as the three layer CCM benchmark can provide a considerable baseline for future studies.

\bibliographystyle{IEEEtran}
\bibliography{refs}

\end{document}